\PassOptionsToPackage{main=english,russian}{babel}
\PassOptionsToPackage{T2A}{fontenc}
\documentclass[onecolumn]{misraj}

\usepackage{lipsum}
\usepackage{pdfpages}
\usepackage{colortbl}
\usepackage{tabulary}
\usepackage{longtable}
\usepackage{array}
\usepackage{placeins}
\usepackage{tablefootnote}
\usepackage{tcolorbox}
\usepackage{wrapfig}
\usepackage{adjustbox}
\usepackage{float}
\usepackage{caption}
\usepackage{breqn} 

\usepackage{pifont}
\usepackage{multirow}
\usepackage{tabularx}

\usepackage[utf8]{inputenc}

\usepackage{arydshln}

\usepackage{xspace} %
\usepackage{makecell} 
\usepackage{multirow}

\usepackage[framemethod=TikZ]{mdframed}
\usepackage{tcolorbox}
\usepackage{multicol}

\usepackage[utf8]{inputenc}
\usepackage[T1]{fontenc}
\usepackage[english]{babel} 
\usepackage{arabtex}
\usepackage{utf8}
\setcode{utf8}

\definecolor{lightblue}{RGB}{211, 227, 252} 
\definecolor{bgblue}{RGB}{247, 250, 255} 

\newcommand*\colourcheck[1]{%
  \expandafter\newcommand\csname #1check\endcsname{\textcolor{#1}{\ding{52}}}%
}
\newcommand*\colourcross[1]{%
  \expandafter\newcommand\csname #1cross\endcsname{\textcolor{#1}{\ding{55}}}%
}
\DeclareSymbolFont{extraup}{U}{zavm}{m}{n}
\DeclareMathSymbol{\vardiamond}{\mathalpha}{extraup}{87}
\colourcheck{blue}
\colourcheck{green}
\colourcheck{red}

\colourcross{blue}
\colourcross{green}
\colourcross{red}
\usepackage{nicematrix}
\usepackage{comment}
\usepackage{amssymb}
\usepackage{arabtex}
\usepackage[utf8]{inputenc}
\usepackage[bottom]{footmisc}
\usepackage{svg}
\usepackage{spverbatim}
\RequirePackage[hidelinks,colorlinks=true,linkcolor=blue,citecolor=blue]{hyperref}
\setcitestyle{number,square}
\definecolor{deeppurple}{HTML}{9e02f7}
\definecolor{forestgreen}{HTML}{2e7d43}

\hypersetup{urlcolor=deeppurple}

\title{Mutarjim: Advancing Bidirectional Arabic-English Translation with a Small Language Model}

\multiauthors
\author{
    name={Khalil Hennara},
    email={hennara@misraj.ai}
}
\author{
    name={Muhammad Hreden},  
    email={hreden@gmail.com}
}

\author{
    name={Mohamed Motasim Hamed},
    email={hamed@misraj.ai}
}
\author{
    name={Zeina Aldallal },
    email={aldallal@misraj.ai}
}
\author{
    name={Sara Chrouf},
    email={sara.chrouf@misraj.ai}
}
\author{
    name={Safwan AlModhayan},
    email={safwan@misraj.ai}
}

\affiliations{
 \includegraphics[width=2cm]{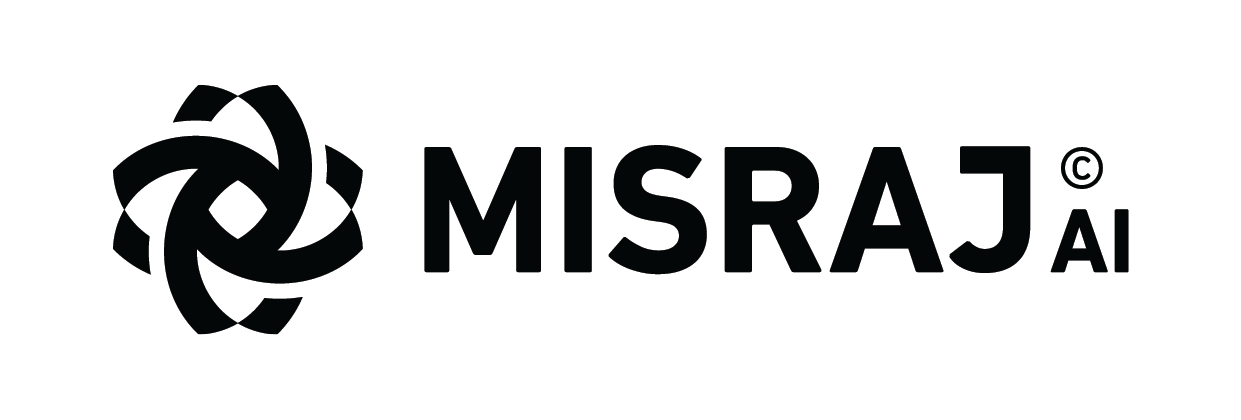}\\
     Khobar, Saudi Arabia \\
     {hennara,hreden,hamed,aldallal,chrouf,safwan}@misraj.ai \\
}
\date{\today}
\abstract{
We introduce \textbf{Mutarjim}, a compact yet powerful language model for bidirectional Arabic-English translation. While large-scale LLMs have shown impressive progress in natural language processing tasks, including machine translation, smaller models. Leveraging this insight, we developed Mutarjim based on \textbf{\textit{Kuwain-1.5B}} \cite{hennara2025kuwain15barabicslm}, a language model tailored for both Arabic and English. Despite its modest size, Mutarjim outperforms much larger models on several established benchmarks, achieved through an optimized two-phase training approach and a carefully curated, high-quality training corpus.. Experimental results show that Mutarjim rivals models up to 20 times larger while significantly reducing computational costs and training requirements. We also introduce \textbf{Tarjama-25}, a new benchmark designed to overcome limitations in existing Arabic-English benchmarking datasets, such as domain narrowness, short sentence lengths, and English-source bias. Tarjama-25 comprises 5,000 expert-reviewed sentence pairs and spans a wide range of domains, offering a more comprehensive and balanced evaluation framework. Notably, Mutarjim achieves state-of-the-art performance on the English-to-Arabic task in Tarjama-25, surpassing even significantly larger and proprietary models like GPT-4o mini. We publicly release Tarjama-25 to support future research and advance the evaluation of Arabic-English translation systems.}

\begin{document}

\renewcommand{\thefootnote}{\fnsymbol{footnote}}

\section{Introduction}
\label{sec:introduction}

Machine translation (MT), a core task in natural language processing (NLP), has made great progress with the rise of Large Language Models (LLMs). However, Arabic machine translation (AMT) is considered a big challenge due to many factors and characteristics of the Arabic language, such as grammar and morphology complexity. Lexical, syntactic, and semantic problems arise when translating the meaning of Arabic words into English and vice versa ~\cite{Survey}.  Despite recent advances in NLP, the Arabic language still lags behind other high-resource languages in terms of translation quality. Existing Arabic-English systems are either limited in their capabilities or are part of larger multilingual models that, while capable of handling many languages, often underperform on Arabic-specific tasks. These models are also computationally demanding, limiting their practicality in low-resource or real-time settings. Consequently, there is increasing interest in developing smaller, task-specific models that balance performance with efficiency while effectively modeling Arabic’s linguistic complexity.

\begin{figure}[ht]
    \centering
    \begin{subfigure}[b]{0.45\textwidth}
        \includegraphics[width=\textwidth]{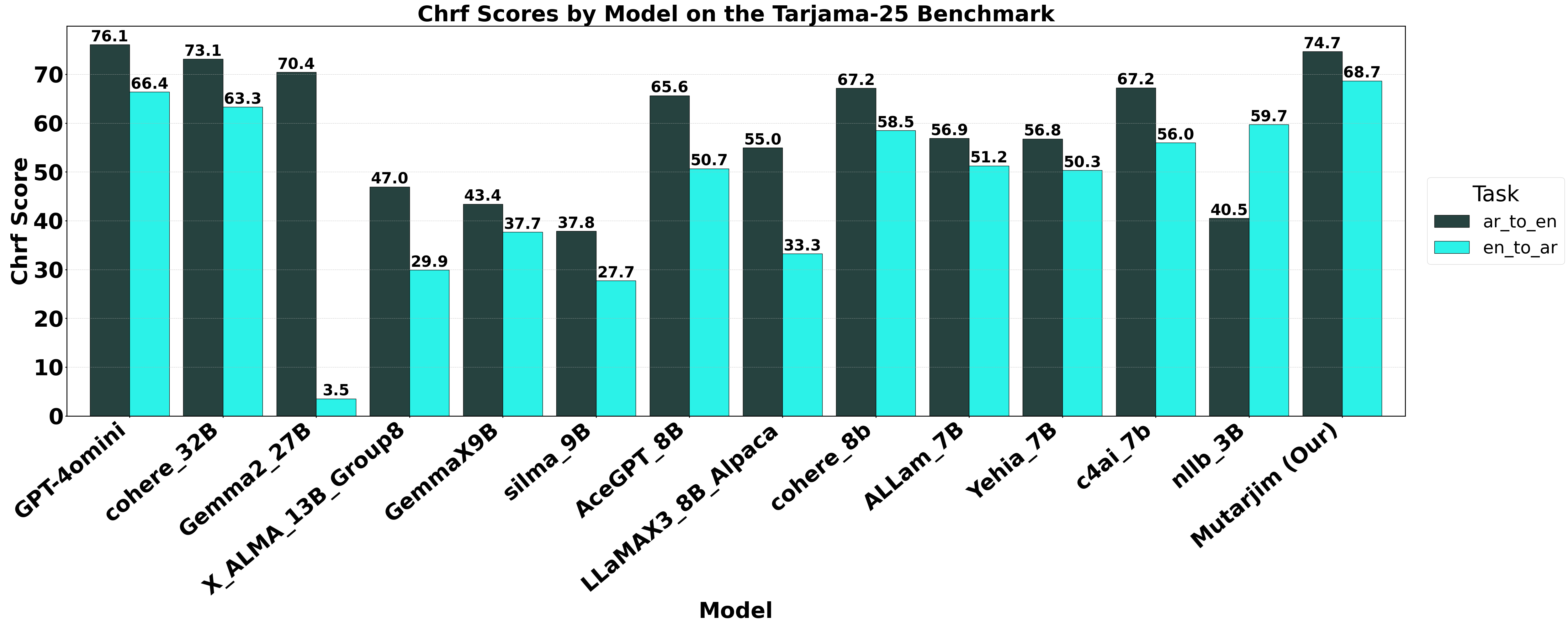}
        \caption*{ChrF++ scores}
    \end{subfigure}
    \hfill
    \begin{subfigure}[b]{0.45\textwidth}
        \includegraphics[width=\textwidth]{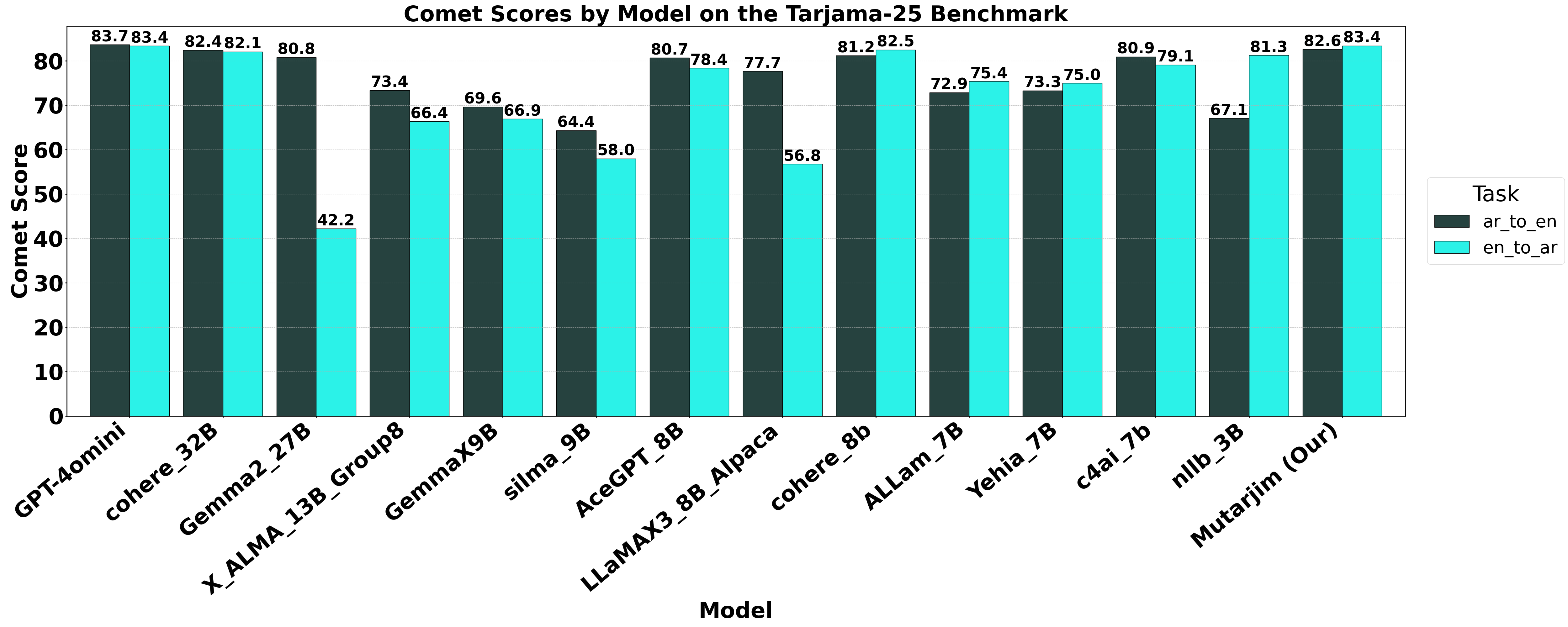}
        \caption*{COMET scores}
    \end{subfigure}
    \caption{Performance of various models on Tarjama-25 sorted by model size, our newly introduced benchmark for Arabic-English translation, evaluated using two metrics: \textbf{ChrF++} (left) and \textbf{COMET} (right).}
    \label{fig:tarjama25_metrics_comparison}
\end{figure}

In this paper, we introduce \textbf{Mutarjim}, a task-specific small language model optimized for Arabic-English translation. Mutarjim is built on Kuwain-1.5B \cite{hennara2025kuwain15barabicslm}, an Arabic-centric decoder-only model.  Mutarjim is trained in two stages: a translation-oriented large-scale pre-training phase and a targeted fine-tuning stage using high-quality parallel corpora. This tailored training approach enables Mutarjim to deliver competitive translation quality and faster inference times. In benchmark evaluations, Mutarjim outperforms models with more than 30 billion parameters, including proprietary systems like GPT-4o mini in both accuracy and efficiency, as shown in Figure~\ref{fig:tarjama25_metrics_comparison}.

To facilitate robust evaluation and future research, we also present \textbf{Tarjama-25}, a new benchmark dataset for bidirectional Arabic-English translation.
Tarjama-25 addresses key limitations of existing datasets, such as short sentence length, English source bias, and limited domain diversity. It comprises 5,000 pairs of expert-curated sentences from diverse domains, with an equal number of examples sourced from Arabic and English originals. All pairs are used to evaluate both Arabic-to-English and English-to-Arabic translation, providing a comprehensive and realistic benchmark for bidirectional translation performance.

Our contributions can be summarized as follows:
\begin{itemize}
    \item We introduce \textbf{Mutarjim}, a compact but powerful decoder-only model specifically optimized for Arabic-English translation.
    \item We present \textbf{Tarjama-25}, a new benchmark for Arabic-English translation, featuring:     \begin{itemize}
        \item Longer and more natural sentence structures;
        \item Balanced translation directionality, with an equal number of source texts originally written in Arabic and English;
        \item Broad domain coverage, spanning general, medical, legal, technological, and other fields;
        \item Careful curation to eliminate contamination from large-scale pre-training corpora, ensuring fair and unbiased evaluation.
        \item Expert-reviewed and human-corrected translations to ensure high linguistic quality and fidelity.
    \end{itemize}
    \item We perform extensive evaluations using multiple standard benchmarks, including 
    WMT24++ \cite{wmt24}, IWSLT2017 \cite{IWSLT2017}, and our newly introduced \textit{Tarjama-25},  and compare \textbf{Mutarjim} against a range of open-source and proprietary models, using automatic metrics BLEU, chrF++, and COMET~\cite{rei2020comet}.
    \item We publicly release both the \textit{Tarjama-25} \footnote{\url{https://huggingface.co/datasets/Misraj/Tarjama-25}} benchmark and its accompanying \textit{evaluation toolkit} \footnote{\url{https://github.com/misraj-ai/Mutarjim-evaluation}} as open-source resources to promote transparency, reproducibility, and further progress in Arabic machine translation research.
\end{itemize}

The rest of the paper is structured as follows: Section~\ref{sec:background} reviews the related works and theoretical foundations relevant to our study. Section~\ref{sec:data} describes the dataset creation process in the two phases. Section~\ref{sec:benchmark} outlines our benchmarking setup.  Section~\ref{sec:method} introduces Mutarjim model with the training methodology. Section~\ref{sec:experiments} details our experiments and results. 
Section~\ref{sec:evaluation} explains our evaluation strategy. Finally, Section~\ref{sec:conclusion} concludes the paper and suggests directions for future research.

\section{Background}
\label{sec:background}

Machine translation has undergone significant evolution, progressing from rule-based approaches necessitating extensive development and maintenance toward statistical and neural paradigms. The development of encoder-decoder and, more recently, decoder-only architectures has played a pivotal role in advancing the field. In this section, we review the major contributions that form the foundation of modern neural MT, focusing on models relevant to Arabic and multilingual translation.

\subsection{Encoder-Decoder Models}

Encoder-decoder Transformer models form the basis of modern neural machine translation (NMT) systems, evolving from general-purpose architectures to specialized, multilingual frameworks. These models vary in scale, language coverage, and specialization, balancing broad applicability with performance in specific languages such as Arabic. This section reviews key models, highlighting their contributions and limitations, particularly in the context of multilingual and Arabic-focused translation. 

The Text-to-Text Transfer Transformer (T5) \cite{t5} introduced a unified text-to-text framework, trained on the English-only C4 dataset with model sizes ranging from 60M to 11B parameters. Although not designed for translation, the flexible architecture of T5 laid the groundwork for subsequent multilingual models. Its primary limitation, the lack of multilingual support, restricts its direct applicability to NMT tasks. Based on T5, mT5 \cite{mT5} extends the text-to-text framework to 101 languages using the multilingual mC4 dataset. However, Arabic constitutes only 1.66\% of its pre-training data, constraining its effectiveness for Arabic translation. Despite strong zero-shot transfer capabilities, mT5 struggles with language-specific nuances, prompting further specialization in later models.
Similarly, Aya-101 \cite{aya101} adapts the mT5-XXL model (13B parameters) using instruction tuning in 100+ languages. This broad multilingual tuning improves the model's ability to translate languages with limited training data, known as low-resource languages. However, its large size results in high training and inference costs, making it less practical for deployment compared to smaller, more specialized models.
In particular, mBART \cite{liu2020multilingual} is a 680M parameter model trained using multilingual denoising pre-training and includes 2.87 billion Arabic tokens in its corpus.
To address broader language coverage, NLLB-200 \cite{costa2022no} is a 3.3B parameter dense model trained on over 18 billion sentence pairs spanning 200 languages. It achieves strong translation performance, particularly in low-resource settings, and is known for its relative robustness against hallucinations. However, its effectiveness diminishes in domain-specific texts such as Islamic or medical content, where it struggles to maintain accuracy and relevance.
In response to the need for Arabic-specific solutions, TURJUMAN \cite{turjuman}, built directly on AraT5 \cite{nagoudi2021arat5},  provides a toolkit for translating 20 source languages into MSA.
\subsection{Decoder-Only Models}

The recent shift toward decoder-only language models has reshaped the machine translation landscape, particularly through their use in autoregressive generation. Unlike traditional encoder-decoder architectures, decoder-only models handle both the source and target text within a single sequence and rely on large-scale pre-training.

A major advantage of decoder-only models lies in their unified architecture for both understanding and generation tasks, enabling efficient transfer and scalability. Prompt-based translation using large models such as GPT-4 \cite{achiam2023gpt} has shown promising results, especially when effective prompting strategies are applied \cite{he2024exploring, lu2023chain, xi2022few, zhu2023multilingual, agrawal2022context, vilar2022prompting}. These approaches often involve feeding a few translation examples directly into the prompt, allowing the model to generalize with minimal supervision.
Beyond prompting, compact decoder-only models have gained traction due to their efficiency. For example, BigTranslate \cite{yang2305bigtranslate} and PARADIGM \cite{xu2023paradigm} explore fine-tuning small open-source LLMs using parallel corpora, achieving competitive translation performance at a fraction of computational cost.
Models like Tower \cite{towerinstruct} and GemmaX2-28 \cite{gemmaX} employ two-phase training: multilingual pre-training followed by fine-tuning conducted according to instructions, enabling domain-specific translation capabilities. Multilingual decoder-only models, such as XALMA \cite{xalma}, scale this approach further, incorporating 50 languages and using MoE layers and adapter modules to manage multilingual transfer.

Within the Arabic domain, general purpose decoder-only models, such as Allam \cite{allam}, AceGPT \cite{acegpt}, and Silma \cite{silma}, are trained primarily in Arabic-centric corpora with objectives to cover various downstream tasks. However, these models are typically not specialized in translation and often lack the fine-grained bilingual alignment necessary for a high-quality Arabic-English translation. Specialized Arabic decoder-only translation models have also emerged. One such model is Lahjawi \cite{hamed2025lahjawi}, a cross-dialect translation system that demonstrated strong performance in both MSA and dialectal Arabic. Lahjawi is specifically undergoing targeted fine-tuning for cross-dialect translation.

Given the trade-off between translation quality, inference speed, and resource efficiency, in this work, we adopt the decoder-only paradigm. \textbf{Mutarjim} follows a two-stage training strategy: large-scale monolingual pre-training followed by supervised fine-tuning using high-quality Arabic-English parallel data. Our choice reflects a growing shift toward compact, efficient, and Arabic-optimized decoder-only systems for machine translation.

\section{Data}
\label{sec:data}
Our training data consists exclusively of bilingual Arabic-English corpora, combining proprietary and open-source sources. For pre-training, we utilize large-scale, domain-diverse parallel datasets to expose the model to a wide range of linguistic patterns and translation contexts.  Fine-tuning focuses on higher-quality data, combining carefully filtered open-source corpora with proprietary datasets curated for accuracy, fluency, and domain relevance. This composition ensures broad coverage during pre-training while emphasizing translation accuracy, fluency, and domain specificity during fine-tuning. 

\subsection{Pre-training Data}

Our pre-training corpus comprises approximately \textbf{10 billion tokens} of bilingual Arabic-English data, used to continue pre-training \textit{kuwain-1.5} the base model and improve its performance in translation tasks. Data are sourced from the OPUS platform~\cite{opus2016}, where we exclusively select Arabic-English parallel corpora, supplemented with proprietary datasets curated internally to improve domain diversity and coverage.

To improve data quality, we applied a set of pre-processing steps, including the removal of sentence pairs with fewer than three tokens, as such samples often lack meaningful context. We also filtered out misaligned examples in which the target sentence is not in the correct language. Pairs with substantial length mismatches are excluded to reduce the risk of partial or noisy translations. Finally, we performed deduplication to eliminate repeated sentence pairs and reduce redundancy in the training data. These filtering steps help maintain a reasonably clean and consistent corpus for pre-training.

\subsection{Fine-tuning Data}

The fine-tuning procedure utilized a precisely selected corpus of approximately \textbf{6 million} Arabic-English parallel sentence pairs. This corpus exhibits substantial diversity and underwent careful filtering procedures to ensure high translation fidelity. The dataset was derived from two principal sources:
\begin{itemize}
\item One portion of the data incorporates translations originally produced by a state-of-the-art LLM. Subsequently, expert linguists inspected a representative subset of these outputs to confirm their accuracy and fluency. To promote stronger Arabic fluency in the model, we emphasized Arabic-centric samples, where Arabic is the source language, at a 2:1 ratio compared to English-centric samples. This approach not only enhances the model's ability to understand and generate Arabic text but also helps preserve the cultural and linguistic richness of the language.
\item The remaining portion comprises high-quality filtered data from OPUS. We applied a combination of automatic and manual review processes, including human inspection of representative subsets. Datasets exhibiting recurring issues such as out-of-context sentences, hallucinations, misinformation, or poor alignment were excluded to maintain overall data integrity.
\end{itemize}

The fine-tuning dataset was designed to align with the domain categories introduced in our benchmark (Section~\ref{sec:benchmark}), ensuring broad and realistic coverage across cultural, legal, scientific, healthcare, religious, and technical domains. We prioritized the inclusion of authentic Arabic source material and maintained a balanced representation across domains to mitigate distributional bias. This targeted curation enables the model to generalize effectively across diverse topics while preserving high translation quality.

\section{Tarjama-25: Bidirectional Arabic-English Translation Benchmark}
\label{sec:benchmark}

\subsection{Motivation and Development}
\begin{figure}[ht]
    \centering
    \includegraphics[width=0.4\textwidth]{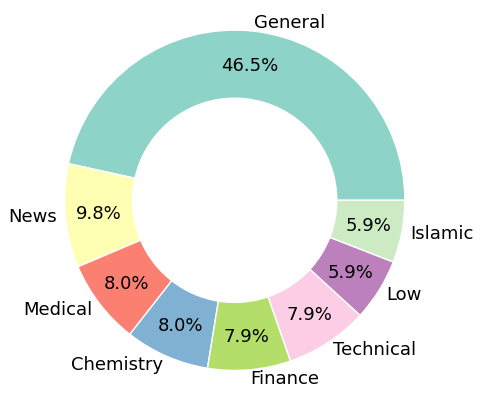}
    \caption{Domain Coverage in Tarjama-25 Benchmark}
\label{fig:tarjama25_domain_distribution}
\end{figure}

Modern MT systems face persistent evaluation challenges due to the limitations of existing benchmarks. To address these gaps, we introduce \textbf{Tarjama-25}, a comprehensive benchmark specifically designed for both Arabic–to–English and English–to–Arabic translation tasks. The current landscape of translation evaluation reveals several critical shortcomings: most publicly available datasets are English-centric (i.e., English is the source language), lacking authentic bidirectional content; benchmarks tend to contain predominantly short sentences (typically 6–30 words), which underutilizes the capacity of modern language models designed to process substantially longer input sequences;  and domain-specific coverage remains limited. Furthermore, potential data contamination from web-scale pre-training and insufficient representation of language-specific characteristics, particularly for Arabic texts, pose additional challenges. To address these challenges, we developed \textbf{Tarjama-25} through a comprehensive data collection and validation pipeline:

\begin{itemize}
    \item We began by collecting 30,000 sentences from authentic Arabic and English sources, each ranging from 50 to 100 words long, ensuring broad domain coverage across scientific, technical, healthcare, cultural, and general interest topics. Half of the data was originally written in Arabic, and the other half in English.
    \item The 30,000 sentences were initially translated using state-of-the-art machine translation systems to create parallel sentence pairs.
    \item From these, 5,000 pairs of sentences were selected for detailed human refinement. Professional translators reviewed and corrected each selected pair to ensure linguistic accuracy and fluency. The final selection maintains a balanced distribution in all domains (Figure~\ref{fig:tarjama25_domain_distribution}).
    \item Finally, domain experts conducted an additional review to validate the accuracy and contextual relevance of the translations within their respective fields.
\end{itemize}

This careful multi-stage process ensures high-quality, human-validated translations with a balanced source language distribution and rich domain diversity, making Tarjama-25 a robust and realistic benchmark for bidirectional Arabic-English translation evaluation.

\subsection{Findings and Recommendations}
Tarjama-25 distinguishes itself through authentic source content in both languages, diverse text lengths, extensive domain coverage, and a strong focus on language-specific subtleties. Our preliminary evaluations reveal that many current MT models, despite their strong performance on existing benchmarks, face significant challenges with Tarjama-25. Detailed evaluation results are presented in Section \ref{sec:evaluation}.

Based on our findings, we recommend:
\begin{enumerate}
\item Development of language-specific authentic benchmarks;
\item Greater emphasis on domain-specific translation capability;
\item Integration of cultural and linguistic nuances in evaluation metrics;
\item Regular benchmark updates to reflect evolving language use.
\end{enumerate}

\section{Method}
\label{sec:method}
For \textbf{Mutarjim}, we build on \textbf{Kuwain-1.5B}~\cite{hennara2025kuwain15barabicslm}, a decoder-only bilingual Arabic-English small language model designed for efficiency in resource-constrained environments. Our approach adopts standard LLM training methodologies commonly used in the field. These methodologies comprise two main phases: pre-training and fine-tuning. To improve translation performance, we introduce targeted modifications within this framework. The pre-training phase is designed to develop a robust bilingual representation, a foundation for the subsequent fine-tuning stage focused specifically on translation tasks.
\begin{figure}[htbp]
    \centering
    \begin{subfigure}[t]{0.45\textwidth} 
        \centering
        \includegraphics[width=\textwidth]{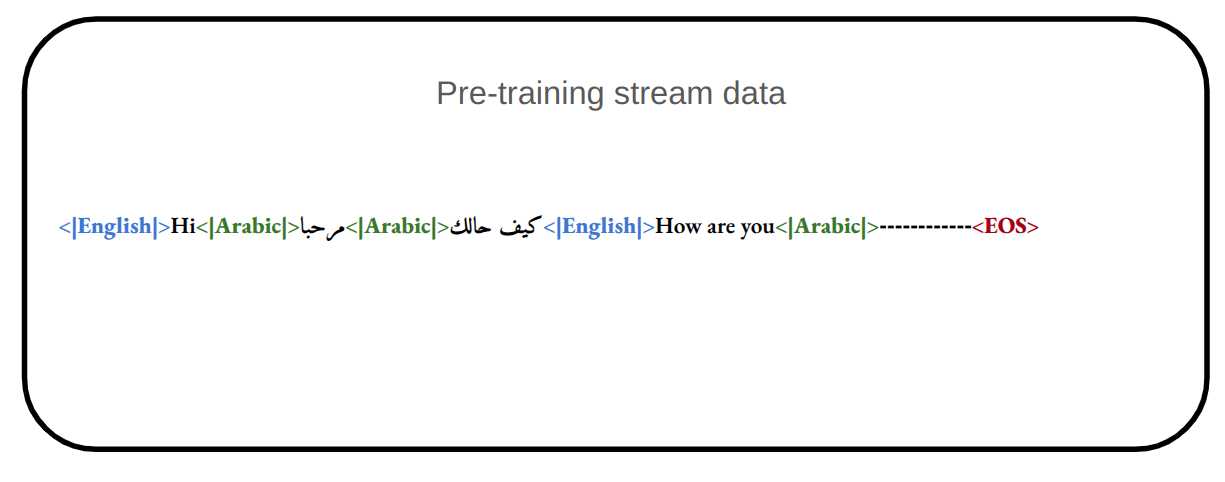} 
        \label{fig:image1}
    \end{subfigure}
    \hfill 
    \begin{subfigure}[t]{0.45\textwidth} 
        \centering
        \includegraphics[width=\textwidth]{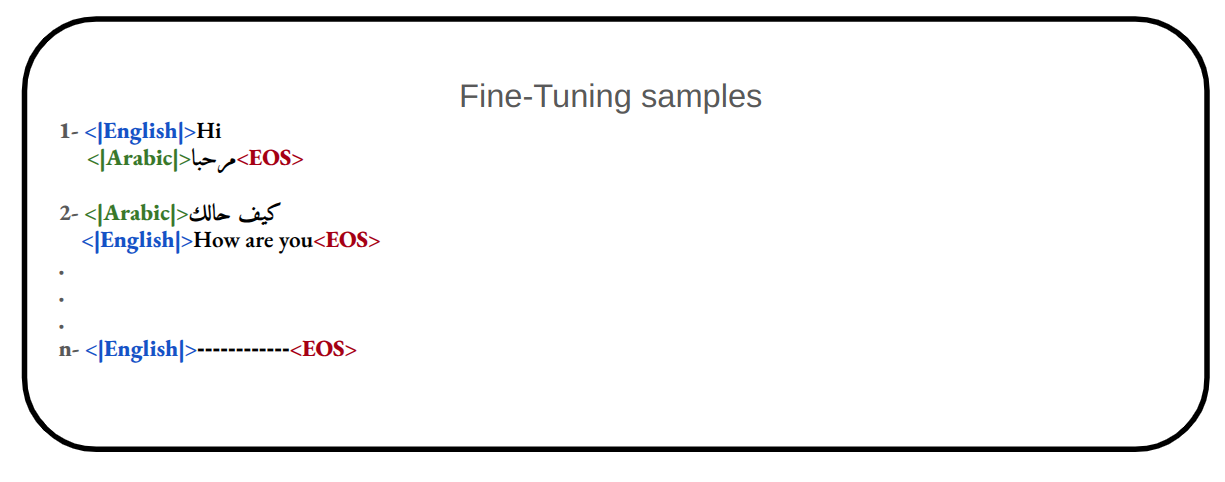} 
        \label{fig:image2}
    \end{subfigure}
    \caption{Illustration of the two data formats used in Mutarjim: (Left) pre-training stream data format; (Right) fine-tuning data sample.}
    \label{fig:side_by_side}
\end{figure}

\subsection{Pre-training Phase} 
\label{subsec:method_pretrain}
Following the successful approaches of recent works such as GemmaX \cite{gemmaX} and Tower \cite{towerinstruct} in continuing pre-training for translation tasks, we further pre-trained our model on English-Arabic parallel data using a next-token prediction objective.

To facilitate the learning process, we introduce two special tokens to our model: \textbf{<|English|>} and \textbf{<|Arabic|>}. We formatted the data as shown on the left side of Figure~\ref{fig:side_by_side}, where English sentences begin with the token <|English|> and Arabic sentences with <|Arabic|>. All pre-training data consist of paired Arabic-English sentences structured according to this format.
During training, the model sees both sentences and is trained to predict the next token over the entire input. To prevent unidirectional translation bias, we randomly select the order of the sentences in each pair. This encourages the model to develop robust bidirectional translation capabilities without favoring a specific source language.

\subsection{Fine-tuning Phase}
\label{subsec:method_finetune}
The fine-tuning phase follows the same format as pre-training, adding a newline between the two sentences for improved structural clarity, as illustrated on the right side of Figure~\ref{fig:side_by_side}. However, unlike the pre-training stage, we apply \textit{causal masking} to the input sentence so that the model is only trained on generating the target sentence from the source, while still using the same next-token prediction objective.

We exclusively use high-quality, human-curated parallel data for this phase to ensure translation accuracy. The model is trained for two epochs over a total of 3 billion tokens, balancing sufficient exposure to high-quality examples with the need to avoid overfitting.
We carefully monitor both training phases to maintain translation quality and prevent performance degradation. Detailed training specifications, including learning rates, batch sizes, and other hyperparameters, are provided in the appendix~\ref{app:B}.

\section{Experiment and Results}
\label{sec:experiments}
To thoroughly evaluate the effectiveness of \textbf{Mutarjim}, we conducted a series of experiments aimed at gaining deeper insights into the challenges and dynamics of Arabic-English translation. Our evaluation focuses on three core aspects. First, we compare unidirectional and bidirectional training setups to assess whether a single model trained in both directions (Arabic–to–English and English–to–Arabic) compromises performance relative to dedicated unidirectional models. Second, we examine the contribution of the continued pre-training phase in enhancing translation quality and improving the model's generalization across domains. Third, we analyze the effect of context length during fine-tuning to understand how sentence length influences performance, particularly when the evaluation samples differ in length from those seen during training. These experiments are conducted using the WMT24++ benchmark, providing a consistent and challenging evaluation framework.

\subsection{Unidirectional vs. Bidirectional Translation Performance}
To assess the impact of directional training on translation quality, we compared unidirectional and bidirectional versions of \textbf{Mutarjim}, focusing on how specializing in a single translation direction affects performance relative to a multitask setup.
We investigated the performance trade-offs between unidirectional models—\textbf{Mutarjim-AR2EN} (Arabic–to–English) and \textbf{Mutarjim-EN2AR} (English–to–Arabic)—and the bidirectional model \textbf{Mutarjim-Bi}. The unidirectional variants were each trained for 3 epochs, while the bidirectional variant was trained for 2 epochs on the combined data. Table~\ref{tab:uni_bid} presents evaluation results using COMET~\cite{rei2020comet} and chrF++ metrics on the WMT24++ benchmark~\cite{wmt24}. Despite being exposed to more diverse data, the bidirectional model showed a slight decrease in performance. Unidirectional models consistently outperformed the bidirectional model, with \textbf{Mutarjim-AR2EN} achieving a COMET score 3.16 points higher than \textbf{Mutarjim-Bi} for Arabic-to-English translation. Ultimately, the choice of model depends on application needs: \textbf{Mutarjim-Bi} offers greater efficiency and flexibility through multitask support, while the unidirectional variants deliver higher translation accuracy for specific directions. Given the compact size of our model (1.5B parameters), the computational cost difference between approaches remains modest.

\begin{table}[ht]
\centering
\small
\begin{tabular}{lccccc}
\toprule
\multirow{2.5}{*}{\textbf{Model}} & \multirow{2.5}{*}{\textbf{Training}} & \multicolumn{2}{c}{\textbf{Arabic → English}} & \multicolumn{2}{c}{\textbf{English → Arabic}} \\
\cmidrule(lr){3-4} \cmidrule(lr){5-6}
 & & COMET & chrF++ & COMET & chrF++ \\
\midrule
Mutarjim-Bi & Bidirectional & 79.73 & 50.27 & 72.86 & 47.04 \\
\addlinespace[0.7ex]
Mutarjim-AR2EN & Unidirectional & \textbf{82.89} & \textbf{54.89} & --- & --- \\
Mutarjim-EN2AR & Unidirectional & --- & --- & \textbf{75.46} & \textbf{48.04} \\ \addlinespace[0.7ex]
\midrule
\end{tabular}
\caption{Performance comparison between bidirectional (Mutarjim-Bi) and unidirectional (Mutarjim-EN2AR or AREN) translation models on WMT24++.}

\label{tab:uni_bid}
\end{table}

\subsection{The Impact of Continued Pre-training Phase}
We evaluated the impact of continued pre-training on translation performance, aiming to determine whether translation-specific pre-training could yield meaningful gains over direct fine-tuning. Although our base model, \textbf{Kuwain}, was initially trained on a substantial corpus, making it a viable candidate for direct fine-tuning, we explore whether targeted continued pre-training on bilingual data enhances downstream translation quality, following recent successes in domain-adaptive pre-training \cite{gemmaX, towerinstruct}.
 
Table~\ref{tab:pre-training} presents a comparison between models trained with and without the additional translation-focused pre-training phase. Models benefiting from this additional phase consistently outperform their counterparts trained solely through fine-tuning, as reflected in both COMET and chrF++ scores.  The gains are evident in both the Arabic–to–English and English–to–Arabic directions, underscoring the general effectiveness of this strategy in translation tasks.
While this approach may not be cost-effective for larger models, it remains computationally feasible for smaller architectures like our 1.5B parameter models. 

\begin{table}[ht]
\centering
\small
\renewcommand{\arraystretch}{1.2}
\begin{tabular}{l|cc|cc}
\hline
\multirow{2}{*}{\textbf{Model}} & \multicolumn{2}{c|}{\textbf{Arabic → English}} & \multicolumn{2}{c}{\textbf{English → Arabic}} \\
 
 & \textbf{COMET} & \textbf{chrF++} & \textbf{COMET} & \textbf{chrF++} \\ 
\hline
\multicolumn{5}{c}{\textit{Without Additional Pre-training}} \\
\hline
Mutarjim-AR2EN & 74.30 & 42.17 & — & — \\ 

Mutarjim-EN2AR & — & — & 61.91 & 34.89 \\ 
\hline
\multicolumn{5}{c}{\textit{With Additional Pre-training}} \\
\hline
Mutarjim-AR2EN & \textbf{82.89} & \textbf{54.89} & — & — \\ 

Mutarjim-EN2AR & — & — & \textbf{75.46} & \textbf{48.04} \\ 
\hline
\end{tabular}
\caption{Effect of additional translation-specific pre-training on model performance, evaluated on WMT24++ test set.}
\label{tab:pre-training}
\end{table}

\subsection{Context Length Effect}
We conducted two independent fine-tuning experiments to evaluate the impact of input length distributions on translation performance. In the first experiment (e1), we fine-tuned the pre-trained Mutarjim model using samples containing more than 30 words, aiming to improve the model’s performance on longer sentences. While this enhanced fluency on long-form content, we observed performance degradation on shorter inputs, with increased hallucinations and irrelevant continuations.

To address this, we performed a second, separate fine-tuning experiment (e2) using the same base model but modifying the training set to include an additional 15\% of short samples (ranging from 2 to 30 words). This experiment sought to balance the model’s ability across varying sequence lengths.
We evaluated both versions on the WMT24++ test set. As shown in Table~\ref{tab:context_eval}, the second experiment (e2) led to improved performance in both directions of translation, confirming the benefit of including shorter sequences in the training data.
\begin{table}[ht]
\centering
\small
\renewcommand{\arraystretch}{1.2}
\begin{tabular}{l|cc|cc}
\hline
\multirow{2}{*}{\textbf{Model}} & \multicolumn{2}{c|}{\textbf{Arabic → English}} & \multicolumn{2}{c}{\textbf{English → Arabic}} \\
 & \textbf{COMET} & \textbf{chrF++} & \textbf{COMET} & \textbf{chrF++} \\ 
\hline
\multicolumn{5}{c}{\textit{Experiment 1 (Long Inputs Only)}} \\
\hline
Mutarjim-AR2EN-e1 & 73.62 & 48.57 & — & — \\ 
Mutarjim-EN2AR-e1 & — & — & 69.07 & 43.40 \\ 
\hline
\multicolumn{5}{c}{\textit{Experiment 2 (Mixed Length Inputs)}} \\
\hline
Mutarjim-AR2EN-e2 & \textbf{74.22} & \textbf{50.84} & — & — \\ 
Mutarjim-EN2AR-e2 & — & — & \textbf{73.56} & \textbf{46.05} \\ 
\hline
\end{tabular}
\caption{Evaluation of models fine-tuned with different input length distributions on the WMT24++ test set.}
\label{tab:context_eval}
\end{table}

\section{Evaluation}
\label{sec:evaluation}
To contextualize the performance of \textbf{Mutarjim}, we compare it against a diverse set of strong decoder-only models that support Arabic and are widely recognized for their translation capabilities. These include general-purpose language models such as AceGPT-8B\cite{acegpt}, ALLam-7B\cite{allam}, C4AI-7B\cite{cohere_for_ai_2024}, Cohere-8B \cite{aya23}, Cohere-32B \cite{aya23}, Gemma2-27B \cite{gemma2}, Silma-9B \cite{silma}, and Yehia-7B \cite{yehia2025}. Furthermore, we include multilingual translation-specialized models such as X-ALMA-13B-Group8 \cite{xalma}, LLaMAX3-8B-Alpaca \cite{llamax}, and GemmaX-9B \cite{gemmaX}. To provide a closer baseline in terms of the architecture and size of the model, we also evaluate against NLLB-3.3B \cite{nllp}, an encoder–decoder model known for its effectiveness in low-resource translation tasks and its widespread adoption in Arabic-English translation.
We evaluated the performance of our model compared to a range of strong baseline models across three established benchmarks: \textbf{WMT24++}, \textbf{IWSLT2017}, and our newly proposed benchmark \textbf{Tarjama-25}. For all benchmarks, we evaluated translation quality using widely adopted metrics, BLEU, chrF++, and COMET, to ensure a comprehensive and fair assessment.

The results for each benchmark are reported in their respective tables: Tarjama-25 in Table~\ref{tab:tarjama25_evaluate}, WMT24++ in Table~\ref{tab:wmt24_evalu}, and IWSLT2017 in Table~\ref{tab:iwslt_evalu}. For consistency, all models are listed in the tables in order of model size. To ensure a fair comparison, we employ model-specific prompts during evaluation, as illustrated in Appendix~\ref{app:C}. To streamline the evaluation pipeline and accelerate inference, we utilize VLLM~\cite{vllm}\footnote{\url{https://docs.vllm.ai/en/stable/}}, which enables efficient batched decoding across decoder-only models.

Although being the smallest among the evaluated models, \textbf{Mutarjim} achieves state-of-the-art performance on the \textbf{Tarjama-25} benchmark for the Arabic-to-English direction in all evaluation metrics, and leads in the English–to–Arabic direction when measured by the BLEU score. It closely trails the much larger \textbf{GPT-4o-mini} model in COMET and chrF++ with only a narrow margin. These results highlight Mutarjim's competitive effectiveness despite its compact size, demonstrating its strength in both translation quality and efficiency. 

Model performance varies noticeably on \textbf{Tarjama-25} compared to existing benchmarks. For example, while \textbf{GPT-4o-mini} excels on WMT24++ and IWSLT2017, its relative performance declines on Tarjama-25. This highlights how standard benchmarks may overlook challenges in domain-specific and bidirectional translation. \textbf{Tarjama-25} helps expose these gaps, offering a more realistic and rigorous assessment of real-world translation capabilities.

Another key observation is the consistent performance gap observed in most models between Arabic-to-English and English-to-Arabic translation, with the former generally yielding better results. This trend is visually illustrated in Figure~\ref{fig:tarjama25_metrics_comparison}, where the disparity, particularly in the chrF++ metric, is pronounced. Several factors may contribute to this asymmetry, including Arabic's rich morphology and syntactic flexibility, which allow for multiple valid translations that current metrics may fail to recognize. Furthermore, the predominance of English-centric training data in many models may hinder their ability to generate fluent and accurate Arabic output.

Notably, Mutarjim demonstrates balanced performance in both translation directions, which we attribute to its Arabic-centric training strategy. This indicates that training with authentic Arabic source data can help mitigate directional bias and improve overall translation fidelity.

\begin{table}[ht]
\fontfamily{ptm}\selectfont
\fontsize{10}{12}\selectfont
\centering
\begin{tabular}{lccccccc}
\toprule
\multirow{2.5}{*}{\textbf{Model}} & \multirow{2.5}{*}{\textbf{Size}} & \multicolumn{3}{c}{\textbf{Arabic → English}} & \multicolumn{3}{c}{\textbf{English → Arabic}} \\
\cmidrule(lr){3-5} \cmidrule(lr){6-8}
& & COMET & Chrf++ & Bleu  & COMET & Chrf++ & Bleu \\ \midrule
Mutarjim & 1.5B & 82.63	& 74.66 & \textbf{55.28} & \textbf{83.41}	& \textbf{68.67} & \textbf{43.71} \\  \addlinespace[0.7ex]
NLLB\cite{no_languge_left_behind}& 3.3B & 67.06	& 40.50 & 24.38 & 81.27	& 59.69 & 30.32 \\  \addlinespace[0.7ex]
c4ai \cite{cohere_for_ai_2024}& 7B & 80.93&	67.24 & 43.34 &79.10&	55.96	&25.18 \\  \addlinespace[0.7ex]
Yehia \cite{yehia2025}& 7B & 73.31 & 56.77 & 32.14 & 74.97 &	50.32	& 20.67  \\  \addlinespace[0.7ex]
ALLam\cite{allam} & 7B & 72.90 & 56.88 & 31.01 & 75.41 & 51.24	& 20.54 \\  \addlinespace[0.7ex]
Cohere \cite{aya23}& 8B & 81.20 & 67.16	& 42.72 & 82.50	& 58.46	& 26.26 \\  \addlinespace[0.7ex]
AceGPT \cite{acegpt} & 8B &  80.71 & 65.63	& 38.67 & 78.39 &50.67	& 20.02 \\  \addlinespace[0.7ex]
LLaMAX3 \cite{llamax}& 8B & 77.72 & 54.95	& 27.86 & 56.76 & 33.25	& 7.63 \\  \addlinespace[0.7ex]
SILMA \cite{silma}& 9B & 64.36 & 37.84 & 15.67 & 58.01 & 27.71	& 5.62 \\  \addlinespace[0.7ex]
GemmaX \cite{gemmaX}& 9B & 69.63 & 43.42	& 19.96 & 66.94 & 37.66 & 9.98 \\  \addlinespace[0.7ex]
XALMA \cite{xalma}& 13B &  73.37 & 46.96 & 21.57 & 66.36 & 29.88	& 6.64 \\  \addlinespace[0.7ex]
Gemma 2 \cite{gemma2}& 27B & 80.81 & 70.42 & 42.78 & 42.20 & 3.52 & 3.08 \\  \addlinespace[0.7ex]
Cohere \cite{aya23}& 32B &  82.44	& 73.10	& 51.16 & 82.09	& 63.29	& 32.25 \\  \addlinespace[0.7ex]
GPT-4o mini \cite{hurst2024gpt}& - & \textbf{83.67} & \textbf{76.08}	& 54.24  & 83.36 & 66.36	& 38.52 \\  \addlinespace[0.7ex] 

\midrule
\end{tabular}
\caption{Performance comparison of bidirectional (Arabic-English) translation models on the \textbf{Tarjama-25} benchmark in terms of COMET, Chrf++, and Bleu.}
\label{tab:tarjama25_evaluate}
\end{table}

\begin{table}[H]
\fontfamily{ptm}\selectfont
\fontsize{10}{12}\selectfont
\centering
\begin{tabular}{lccccccc}
\toprule
\multirow{2.5}{*}{\textbf{Model}} & \multirow{2.5}{*}{\textbf{Size}} & \multicolumn{3}{c}{\textbf{Arabic → English}} & \multicolumn{3}{c}{\textbf{English → Arabic}} \\
\cmidrule(lr){3-5} \cmidrule(lr){6-8}
& & COMET & Chrf++ & Bleu  & COMET & Chrf++ & Bleu \\ \midrule
Mutarjim & 1.5B & 72.99 & 52.27	& 19.26 & 75.46	& 48.04	& 17.99 \\  \addlinespace[0.7ex]
NLLB\cite{no_languge_left_behind}& 3.3B &76.71&	50.13&	25.50& 77.75&	45.89&	16.03 \\ \addlinespace[0.7ex]
c4ai \cite{cohere_for_ai_2024}& 7B & 79.27&54.91&	26.35& 72.45&44.32&	14.19 \\ \addlinespace[0.7ex]
Yehia \cite{yehia2025}& 7B & 72.72&	47.58&	15.39& 72.23&	41.12&	10.69  \\ \addlinespace[0.7ex]
ALLam\cite{allam} & 7B & 72.00&	46.80&	15.01& 71.89&	41.45&	10.41 \\ \addlinespace[0.7ex]
Cohere \cite{aya23}& 8B & 78.89&	54.06&	24.96& 74.80&	44.95&	14.08 \\  \addlinespace[0.7ex]
AceGPT \cite{acegpt} & 8B & 78.18&	52.25&	21.21& 73.65&	40.55&	11.37 \\ \addlinespace[0.7ex]
LLaMAX3 \cite{llamax}& 8B & 75.91&	48.18&	18.89& 57.31&	28.83&	4.03 \\ \addlinespace[0.7ex]
SILMA \cite{silma}& 9B & 71.33&	38.96&	16.44& 60.54&	26.75&	4.97 \\ \addlinespace[0.7ex]
GemmaX \cite{gemmaX}& 9B & 77.82&	50.80&	22.67& 70.21&	38.81&	9.83 \\ \addlinespace[0.7ex]
XALMA \cite{xalma}& 13B & 76.84&	48.65&	19.34& 69.19&	33.23& 7.54 \\ \addlinespace[0.7ex]
Gemma 2 \cite{gemma2}& 27B & 72.79& 51.09&	16.59& 54.00&	32.66&	4.77 \\ \addlinespace[0.7ex]
Cohere \cite{aya23}& 32B & 79.77&	57.05&27.98&  72.74&	47.13&	15.84  \\ \addlinespace[0.7ex]
GPT-4o mini \cite{hurst2024gpt}& - &\textbf{83.29}& \textbf{58.24}&	\textbf{29.23}& \textbf{82.32} & \textbf{50.03} & \textbf{20.48} \\
\midrule
\end{tabular}
\caption{Performance comparison of bidirectional (Arabic-English) translation models on the\textbf{ WMT24++} benchmark in terms of COMET, Chrf++, and Bleu.}
\label{tab:wmt24_evalu}
\end{table}

\begin{table}[H]
\fontfamily{ptm}\selectfont
\fontsize{10}{12}\selectfont
\centering
\begin{tabular}{lccccccc}
\toprule
\multirow{2.5}{*}{\textbf{Model}} & \multirow{2.5}{*}{\textbf{Size}} & \multicolumn{3}{c}{\textbf{Arabic → English}} & \multicolumn{3}{c}{\textbf{English → Arabic}} \\
\cmidrule(lr){3-5} \cmidrule(lr){6-8}
& & COMET & Chrf++ & Bleu  & COMET & Chrf++ & Bleu \\ \midrule
Mutarjim & 1.5B & 82.89	&54.89&	31.00 & 79.76 &	44.21&	12.74  \\  \addlinespace[0.7ex]
NLLB\cite{no_languge_left_behind}& 3.3B &- & - & - & - & - & - \\  \addlinespace[0.7ex]
c4ai \cite{cohere_for_ai_2024}& 7B & 83.99 & 56.64	& 33.64 & 77.41 & 	40.50 &	9.14 \\  \addlinespace[0.7ex]
Yehia \cite{yehia2025}& 7B & 75.58&	47.38&	15.93& 76.22&	38.41&	6.65 \\  \addlinespace[0.7ex]
ALLam\cite{allam} & 7B & 75.64&	37.36&	5.89& 75.25&	46.54&	14.79 \\  \addlinespace[0.7ex]
Cohere \cite{aya23}& 8B & 83.60&	55.83&	31.71& 79.05&	42.36&	9.10 \\  \addlinespace[0.7ex]
AceGPT \cite{acegpt} & 8B & 81.72&	52.83&	26.26& 79.62&	40.23&	9.25 \\  \addlinespace[0.7ex]
LLaMAX3 \cite{llamax}& 8B & 81.04&	49.17&	24.28&67.79&30.17&	4.18 \\  \addlinespace[0.7ex]
SILMA \cite{silma}& 9B & 78.55&47.57&	24.28& 69.59&	30.03&	5.11 \\  \addlinespace[0.7ex]
GemmaX \cite{gemmaX}& 9B & 82.06&	53.30&	30.25& 76.17	& 37.17	& 7.10 \\  \addlinespace[0.7ex]
XALMA \cite{xalma}& 13B & 80.06&	49.04&	24.10& 76.41&	36.99&	7.13 \\  \addlinespace[0.7ex]
Gemma 2 \cite{gemma2}& 27B & -& - & - & 48.56&	22.28&	1.57 \\  \addlinespace[0.7ex]
Cohere \cite{aya23}& 32B & 84.30&	59.02&	35.37& 74.63&	43.53&	8.93 \\  \addlinespace[0.7ex]
GPT-4o mini \cite{hurst2024gpt}& - & \textbf{86.37}& \textbf{60.48} & \textbf{36.86} & \textbf{87.14} & 	\textbf{47.63}&	\textbf{15.50}  \\ 
\midrule
\end{tabular}
\caption{Performance comparison of bidirectional (Arabic-English) translation models on the \textbf{IWSLT-2017}  benchmark in terms of COMET, Chrf++, and Bleu.}
\label{tab:iwslt_evalu}
\end{table}

\section{Conclusion}
\label{sec:conclusion}
In this work, we introduce \textbf{Mutarjim}, an efficient and compact small language model, optimized for bidirectional Arabic-English machine translation while providing rich and accurate output. We also present a new benchmark \textbf{Tarjama-25}, a diverse and representative dataset for bidirectional Arabic-English MT evaluation. Our evaluation and experiments demonstrate that \textbf{Mutarjim} achieves competitive performance against larger models while requiring significantly fewer computational resources. The model's compact architecture enables deployment in resource-constrained environments without sacrificing translation quality. Future work will focus on scaling up the model architecture and training on larger multilingual datasets to support translation between Arabic and multiple languages, including French, Turkish, and Japanese, to create a comprehensive multilingual translation system while maintaining efficiency.

\bibliography{main}

\begin{thebibliography}{38}
\providecommand{\natexlab}[1]{#1}
\providecommand{\url}[1]{\texttt{#1}}
\expandafter\ifx\csname urlstyle\endcsname\relax
  \providecommand{\doi}[1]{doi: #1}\else
  \providecommand{\doi}{doi: \begingroup \urlstyle{rm}\Url}\fi

\bibitem[Achiam et~al.(2023)Achiam, Adler, Agarwal, Ahmad, Akkaya, Aleman, Almeida, Altenschmidt, Altman, Anadkat, et~al.]{achiam2023gpt}
Josh Achiam, Steven Adler, Sandhini Agarwal, Lama Ahmad, Ilge Akkaya, Florencia~Leoni Aleman, Diogo Almeida, Janko Altenschmidt, Sam Altman, Shyamal Anadkat, et~al.
\newblock Gpt-4 technical report.
\newblock \emph{arXiv preprint arXiv:2303.08774}, 2023.

\bibitem[Agrawal et~al.(2022)Agrawal, Zhou, Lewis, Zettlemoyer, and Ghazvininejad]{agrawal2022context}
Sweta Agrawal, Chunting Zhou, Mike Lewis, Luke Zettlemoyer, and Marjan Ghazvininejad.
\newblock In-context examples selection for machine translation.
\newblock \emph{arXiv preprint arXiv:2212.02437}, 2022.

\bibitem[Alves et~al.(2024)Alves, Pombal, Guerreiro, Martins, Alves, Farajian, Peters, Rei, Fernandes, Agrawal, et~al.]{towerinstruct}
Duarte~M Alves, Jos{\'e} Pombal, Nuno~M Guerreiro, Pedro~H Martins, Jo{\~a}o Alves, Amin Farajian, Ben Peters, Ricardo Rei, Patrick Fernandes, Sweta Agrawal, et~al.
\newblock Tower: An open multilingual large language model for translation-related tasks.
\newblock \emph{arXiv preprint arXiv:2402.17733}, 2024.

\bibitem[Aryabumi et~al.(2024)Aryabumi, Dang, Talupuru, Dash, Cairuz, Lin, Venkitesh, Smith, Campos, Tan, et~al.]{aya23}
Viraat Aryabumi, John Dang, Dwarak Talupuru, Saurabh Dash, David Cairuz, Hangyu Lin, Bharat Venkitesh, Madeline Smith, Jon~Ander Campos, Yi~Chern Tan, et~al.
\newblock Aya 23: Open weight releases to further multilingual progress.
\newblock \emph{arXiv preprint arXiv:2405.15032}, 2024.

\bibitem[Baligh \& Mohammed(2022)Baligh and Mohammed]{Survey}
Babaali Baligh and Salem Mohammed.
\newblock Arabic machine translation: A panoramic survey.
\newblock \emph{SSRN Electronic Journal}, 01 2022.
\newblock \doi{10.2139/ssrn.4312742}.

\bibitem[Bari et~al.(2024)Bari, Alnumay, Alzahrani, Alotaibi, Alyahya, AlRashed, Mirza, Alsubaie, Alahmed, Alabduljabbar, et~al.]{allam}
M~Saiful Bari, Yazeed Alnumay, Norah~A Alzahrani, Nouf~M Alotaibi, Hisham~A Alyahya, Sultan AlRashed, Faisal~A Mirza, Shaykhah~Z Alsubaie, Hassan~A Alahmed, Ghadah Alabduljabbar, et~al.
\newblock Allam: Large language models for arabic and english.
\newblock \emph{arXiv preprint arXiv:2407.15390}, 2024.

\bibitem[Cettolo et~al.(2017)Cettolo, Federico, Bentivogli, Niehues, St{\"u}ker, Sudoh, Yoshino, and Federmann]{IWSLT2017}
Mauro Cettolo, Marcello Federico, Luisa Bentivogli, Jan Niehues, Sebastian St{\"u}ker, Katsuhito Sudoh, Koichiro Yoshino, and Christian Federmann.
\newblock Overview of the {IWSLT} 2017 evaluation campaign.
\newblock In \emph{Proceedings of the 14th International Conference on Spoken Language Translation}, pp.\  2--14, Tokyo, Japan, December 14-15 2017. International Workshop on Spoken Language Translation.
\newblock URL \url{https://aclanthology.org/2017.iwslt-1.1}.

\bibitem[{Cohere For AI}(2024)]{cohere_for_ai_2024}
{Cohere For AI}.
\newblock c4ai-command-r-07-arabic-2025, 2024.
\newblock URL \url{https://huggingface.co/CohereForAI/c4ai-command-r-08-2024}.

\bibitem[Costa-Juss{\`a} et~al.(2022{\natexlab{a}})Costa-Juss{\`a}, Cross, {\c{C}}elebi, Elbayad, Heafield, Heffernan, Kalbassi, Lam, Licht, Maillard, et~al.]{costa2022no}
Marta~R Costa-Juss{\`a}, James Cross, Onur {\c{C}}elebi, Maha Elbayad, Kenneth Heafield, Kevin Heffernan, Elahe Kalbassi, Janice Lam, Daniel Licht, Jean Maillard, et~al.
\newblock No language left behind: Scaling human-centered machine translation.
\newblock \emph{arXiv preprint arXiv:2207.04672}, 2022{\natexlab{a}}.

\bibitem[Costa-Juss{\`a} et~al.(2022{\natexlab{b}})Costa-Juss{\`a}, Cross, {\c{C}}elebi, Elbayad, Heafield, Heffernan, Kalbassi, Lam, Licht, Maillard, et~al.]{no_languge_left_behind}
Marta~R Costa-Juss{\`a}, James Cross, Onur {\c{C}}elebi, Maha Elbayad, Kenneth Heafield, Kevin Heffernan, Elahe Kalbassi, Janice Lam, Daniel Licht, Jean Maillard, et~al.
\newblock No language left behind: Scaling human-centered machine translation.
\newblock \emph{arXiv preprint arXiv:2207.04672}, 2022{\natexlab{b}}.

\bibitem[Cui et~al.(2025)Cui, Gao, Liu, Luan, et~al.]{gemmaX}
Menglong Cui, Pengzhi Gao, Wei Liu, Jian Luan, et~al.
\newblock Multilingual machine translation with open large language models at practical scale: An empirical study.
\newblock \emph{arXiv preprint arXiv:2502.02481}, 2025.

\bibitem[Deutsch et~al.(2025)Deutsch, Briakou, Caswell, Finkelstein, Galor, Juraska, Kovacs, Lui, Rei, Riesa, et~al.]{wmt24}
Daniel Deutsch, Eleftheria Briakou, Isaac Caswell, Mara Finkelstein, Rebecca Galor, Juraj Juraska, Geza Kovacs, Alison Lui, Ricardo Rei, Jason Riesa, et~al.
\newblock Wmt24++: Expanding the language coverage of wmt24 to 55 languages \& dialects.
\newblock \emph{arXiv preprint arXiv:2502.12404}, 2025.

\bibitem[Hamed et~al.(2025)Hamed, Hreden, Hennara, Aldallal, Chrouf, and AlModhayan]{hamed2025lahjawi}
Mohamed~Motasim Hamed, Muhammad Hreden, Khalil Hennara, Zeina Aldallal, Sara Chrouf, and Safwan AlModhayan.
\newblock Lahjawi: Arabic cross-dialect translator.
\newblock In \emph{Proceedings of the 4th Workshop on Arabic Corpus Linguistics (WACL-4)}, pp.\  12--24, 2025.

\bibitem[He et~al.(2024)He, Liang, Jiao, Zhang, Yang, Wang, Tu, Shi, and Wang]{he2024exploring}
Zhiwei He, Tian Liang, Wenxiang Jiao, Zhuosheng Zhang, Yujiu Yang, Rui Wang, Zhaopeng Tu, Shuming Shi, and Xing Wang.
\newblock Exploring human-like translation strategy with large language models.
\newblock \emph{Transactions of the Association for Computational Linguistics}, 12:\penalty0 229--246, 2024.

\bibitem[Hennara et~al.(2025)Hennara, Chrouf, Hamed, Aldallal, Hadid, and AlModhayan]{hennara2025kuwain15barabicslm}
Khalil Hennara, Sara Chrouf, Mohamed~Motaism Hamed, Zeina Aldallal, Omar Hadid, and Safwan AlModhayan.
\newblock Kuwain 1.5 b: An arabic slm via language injection.
\newblock \emph{arXiv preprint arXiv:2504.15120}, 2025.

\bibitem[Huang et~al.(2023)Huang, Yu, Zhu, Sun, Cheng, Song, Chen, Alharthi, An, He, et~al.]{acegpt}
Huang Huang, Fei Yu, Jianqing Zhu, Xuening Sun, Hao Cheng, Dingjie Song, Zhihong Chen, Abdulmohsen Alharthi, Bang An, Juncai He, et~al.
\newblock Acegpt, localizing large language models in arabic.
\newblock \emph{arXiv preprint arXiv:2309.12053}, 2023.

\bibitem[Hurst et~al.(2024)Hurst, Lerer, Goucher, Perelman, Ramesh, Clark, Ostrow, Welihinda, Hayes, Radford, et~al.]{hurst2024gpt}
Aaron Hurst, Adam Lerer, Adam~P Goucher, Adam Perelman, Aditya Ramesh, Aidan Clark, AJ~Ostrow, Akila Welihinda, Alan Hayes, Alec Radford, et~al.
\newblock Gpt-4o system card.
\newblock \emph{arXiv preprint arXiv:2410.21276}, 2024.

\bibitem[Kwon et~al.(2023)Kwon, Li, Zhuang, Sheng, Zheng, Yu, Gonzalez, Zhang, and Stoica]{vllm}
Woosuk Kwon, Zhuohan Li, Siyuan Zhuang, Ying Sheng, Lianmin Zheng, Cody~Hao Yu, Joseph~E. Gonzalez, Hao Zhang, and Ion Stoica.
\newblock Efficient memory management for large language model serving with pagedattention.
\newblock In \emph{Proceedings of the ACM SIGOPS 29th Symposium on Operating Systems Principles}, 2023.

\bibitem[Liu et~al.(2020)Liu, Gu, Goyal, Li, Edunov, Ghazvininejad, Lewis, and Zettlemoyer]{liu2020multilingual}
Yinhan Liu, Jiatao Gu, Naman Goyal, Xian Li, Sergey Edunov, Marjan Ghazvininejad, Mike Lewis, and Luke Zettlemoyer.
\newblock Multilingual denoising pre-training for neural machine translation.
\newblock \emph{Transactions of the Association for Computational Linguistics}, 8:\penalty0 726--742, 2020.

\bibitem[Lu et~al.(2023)Lu, Yang, Huang, Zhang, Lam, and Wei]{lu2023chain}
Hongyuan Lu, Haoran Yang, Haoyang Huang, Dongdong Zhang, Wai Lam, and Furu Wei.
\newblock Chain-of-dictionary prompting elicits translation in large language models. arxiv e-prints, page.
\newblock \emph{arXiv preprint arXiv:2305.06575}, 2023.

\bibitem[Lu et~al.(2024)Lu, Zhu, Li, Qiao, and Yuan]{llamax}
Yinquan Lu, Wenhao Zhu, Lei Li, Yu~Qiao, and Fei Yuan.
\newblock Llamax: Scaling linguistic horizons of llm by enhancing translation capabilities beyond 100 languages.
\newblock \emph{arXiv preprint arXiv:2407.05975}, 2024.

\bibitem[Nagoudi et~al.(2021)Nagoudi, Elmadany, and Abdul-Mageed]{nagoudi2021arat5}
El~Moatez~Billah Nagoudi, AbdelRahim Elmadany, and Muhammad Abdul-Mageed.
\newblock Arat5: Text-to-text transformers for arabic language generation.
\newblock \emph{arXiv preprint arXiv:2109.12068}, 2021.

\bibitem[Nagoudi et~al.(2022)Nagoudi, Elmadany, and Abdul-Mageed]{turjuman}
El~Moatez~Billah Nagoudi, AbdelRahim Elmadany, and Muhammad Abdul-Mageed.
\newblock Turjuman: A public toolkit for neural arabic machine translation, 2022.
\newblock URL \url{https://arxiv.org/abs/2206.03933}.

\bibitem[Navid-AI(2025)]{yehia2025}
Navid-AI.
\newblock Yehia 7b preview.
\newblock \url{https://huggingface.co/Navid-AI/Yehia-7B-preview}, 2025.

\bibitem[Raffel et~al.(2023)Raffel, Shazeer, Roberts, Lee, Narang, Matena, Zhou, Li, and Liu]{t5}
Colin Raffel, Noam Shazeer, Adam Roberts, Katherine Lee, Sharan Narang, Michael Matena, Yanqi Zhou, Wei Li, and Peter~J. Liu.
\newblock Exploring the limits of transfer learning with a unified text-to-text transformer, 2023.
\newblock URL \url{https://arxiv.org/abs/1910.10683}.

\bibitem[Rei et~al.(2020)Rei, Stewart, Farinha, and Lavie]{rei2020comet}
Ricardo Rei, Craig Stewart, Ana~C Farinha, and Alon Lavie.
\newblock Comet: A neural framework for mt evaluation.
\newblock \emph{arXiv preprint arXiv:2009.09025}, 2020.

\bibitem[Team et~al.(2024)Team, Riviere, Pathak, Sessa, Hardin, Bhupatiraju, Hussenot, Mesnard, Shahriari, Ram{\'e}, et~al.]{gemma2}
Gemma Team, Morgane Riviere, Shreya Pathak, Pier~Giuseppe Sessa, Cassidy Hardin, Surya Bhupatiraju, L{\'e}onard Hussenot, Thomas Mesnard, Bobak Shahriari, Alexandre Ram{\'e}, et~al.
\newblock Gemma 2: Improving open language models at a practical size.
\newblock \emph{arXiv preprint arXiv:2408.00118}, 2024.

\bibitem[Team et~al.(2022)Team, Costa-jussà, Cross, Çelebi, Elbayad, Heafield, Heffernan, Kalbassi, Lam, Licht, Maillard, Sun, Wang, Wenzek, Youngblood, Akula, Barrault, Gonzalez, Hansanti, Hoffman, Jarrett, Sadagopan, Rowe, Spruit, Tran, Andrews, Ayan, Bhosale, Edunov, Fan, Gao, Goswami, Guzmán, Koehn, Mourachko, Ropers, Saleem, Schwenk, and Wang]{nllp}
NLLB Team, Marta~R. Costa-jussà, James Cross, Onur Çelebi, Maha Elbayad, Kenneth Heafield, Kevin Heffernan, Elahe Kalbassi, Janice Lam, Daniel Licht, Jean Maillard, Anna Sun, Skyler Wang, Guillaume Wenzek, Al~Youngblood, Bapi Akula, Loic Barrault, Gabriel~Mejia Gonzalez, Prangthip Hansanti, John Hoffman, Semarley Jarrett, Kaushik~Ram Sadagopan, Dirk Rowe, Shannon Spruit, Chau Tran, Pierre Andrews, Necip~Fazil Ayan, Shruti Bhosale, Sergey Edunov, Angela Fan, Cynthia Gao, Vedanuj Goswami, Francisco Guzmán, Philipp Koehn, Alexandre Mourachko, Christophe Ropers, Safiyyah Saleem, Holger Schwenk, and Jeff Wang.
\newblock No language left behind: Scaling human-centered machine translation, 2022.
\newblock URL \url{https://arxiv.org/abs/2207.04672}.

\bibitem[Team(2024)]{silma}
Silma Team.
\newblock Silma.
\newblock 2024.
\newblock URL \url{https://www.silma.ai}.

\bibitem[Tiedemann(2016)]{opus2016}
J{\"o}rg Tiedemann.
\newblock {OPUS} {--} parallel corpora for everyone.
\newblock In \emph{Proceedings of the 19th Annual Conference of the European Association for Machine Translation: Projects/Products}, Riga, Latvia, May 30{--}June 1 2016. Baltic Journal of Modern Computing.
\newblock URL \url{https://aclanthology.org/2016.eamt-2.8}.

\bibitem[{\"U}st{\"u}n et~al.(2024){\"U}st{\"u}n, Aryabumi, Yong, Ko, D'souza, Onilude, Bhandari, Singh, Ooi, Kayid, et~al.]{aya101}
Ahmet {\"U}st{\"u}n, Viraat Aryabumi, Zheng-Xin Yong, Wei-Yin Ko, Daniel D'souza, Gbemileke Onilude, Neel Bhandari, Shivalika Singh, Hui-Lee Ooi, Amr Kayid, et~al.
\newblock Aya model: An instruction finetuned open-access multilingual language model.
\newblock \emph{arXiv preprint arXiv:2402.07827}, 2024.

\bibitem[Vilar et~al.(2022)Vilar, Freitag, Cherry, Luo, Ratnakar, and Foster]{vilar2022prompting}
David Vilar, Markus Freitag, Colin Cherry, Jiaming Luo, Viresh Ratnakar, and George Foster.
\newblock Prompting palm for translation: Assessing strategies and performance.
\newblock \emph{arXiv preprint arXiv:2211.09102}, 2022.

\bibitem[Xi et~al.(2022)Xi, Mihaylov, Artetxe, Wang, Chen, Simig, Ott, Goyal, Bhosale, Jingfei, et~al.]{xi2022few}
Victoria~Lin Xi, Todor Mihaylov, Mikel Artetxe, Tianlu Wang, Shuohui Chen, Daniel Simig, Myle Ott, Naman Goyal, Shruti Bhosale, Du~Jingfei, et~al.
\newblock Few-shot learning with multilingual generative language models.
\newblock In \emph{Proceedings of the 2022 Conference on Empirical Methods in Natural Language Processing}, pp.\  9019--9052. Association for Computational Linguistics Abu Dhabi, United Arab Emirates, 2022.

\bibitem[Xu et~al.(2023)Xu, Kim, Sharaf, and Awadalla]{xu2023paradigm}
Haoran Xu, Young~Jin Kim, Amr Sharaf, and Hany~Hassan Awadalla.
\newblock A paradigm shift in machine translation: Boosting translation performance of large language models.
\newblock \emph{arXiv preprint arXiv:2309.11674}, 2023.

\bibitem[Xu et~al.(2024)Xu, Murray, Koehn, Hoang, Eriguchi, and Khayrallah]{xalma}
Haoran Xu, Kenton Murray, Philipp Koehn, Hieu Hoang, Akiko Eriguchi, and Huda Khayrallah.
\newblock X-alma: Plug \& play modules and adaptive rejection for quality translation at scale.
\newblock \emph{arXiv preprint arXiv:2410.03115}, 2024.

\bibitem[Xue et~al.(2021)Xue, Constant, Roberts, Kale, Al-Rfou, Siddhant, Barua, and Raffel]{mT5}
Linting Xue, Noah Constant, Adam Roberts, Mihir Kale, Rami Al-Rfou, Aditya Siddhant, Aditya Barua, and Colin Raffel.
\newblock mt5: A massively multilingual pre-trained text-to-text transformer, 2021.
\newblock URL \url{https://arxiv.org/abs/2010.11934}.

\bibitem[Yang et~al.(2023)Yang, Li, Zhang, and Zong]{yang2305bigtranslate}
W~Yang, C~Li, J~Zhang, and C~Zong.
\newblock Bigtranslate: Augmenting large language models with multilingual translation capability over 100 languages. arxiv 2023.
\newblock \emph{arXiv preprint arXiv:2305.18098}, 2023.

\bibitem[Zhu et~al.(2023)Zhu, Liu, Dong, Xu, Huang, Kong, Chen, and Li]{zhu2023multilingual}
Wenhao Zhu, Hongyi Liu, Qingxiu Dong, Jingjing Xu, Shujian Huang, Lingpeng Kong, Jiajun Chen, and Lei Li.
\newblock Multilingual machine translation with large language models: Empirical results and analysis.
\newblock \emph{arXiv preprint arXiv:2304.04675}, 2023.

\end{thebibliography}

\clearpage
\newpage
\appendix

\section{Evaluation details}

\section{Traning Details}
\label{app:B}
We trained Mutarjim using a two-stage approach (pre-training and fine-tuning) on 8 NVIDIA H100 GPUs. Table \ref{tab:hyperparameters} summarizes the key hyperparameters for both phases.

\begin{table}[H]
\fontfamily{ptm}\selectfont
\fontsize{10}{12}\selectfont
\small
\centering
\begin{tabular}{lcc}
\toprule
\textbf{Hyperparameter} & \textbf{Pre-training} & \textbf{Fine-tuning} \\
\midrule
Max Learning Rate & $1 \times 10^{-4}$ & $8 \times 10^{-5}$ \\
Learning Rate Schedule & Cosine  & Cosine \\
Weight Decay & $0.1$ & $0.01$ \\
Optimizer & AdamW & AdamW \\
Batch Size & $1024$ & $4096$ \\
Training Steps & $4$K & $3.3$K \\
Context Length & $2048$ & $512$ \\
GPUs & $8 \times H100$ & $8 \times H100$ \\
\midrule
\end{tabular}
\caption{Training Hyperparameters}
\label{tab:hyperparameters}
\end{table}

\section{Mutarjim Translation Examples}\label{secd}
Tables~\ref{tab:e2a_translations} and~\ref{tab:a2e_translations} present qualitative examples of Mutarjim’s performance in both translation directions. These examples cover a range of domains—including mathematics, structured data, biomedical content, and informal discourse—and illustrate the model’s ability to produce accurate, fluent translations that preserve both meaning and structure.
\begin{table}[ht]
    \centering
    \begin{tabular}{c}
     \includegraphics[
     clip,
     trim=0cm 19.6cm 0cm 2cm, 
     width=1\textwidth]{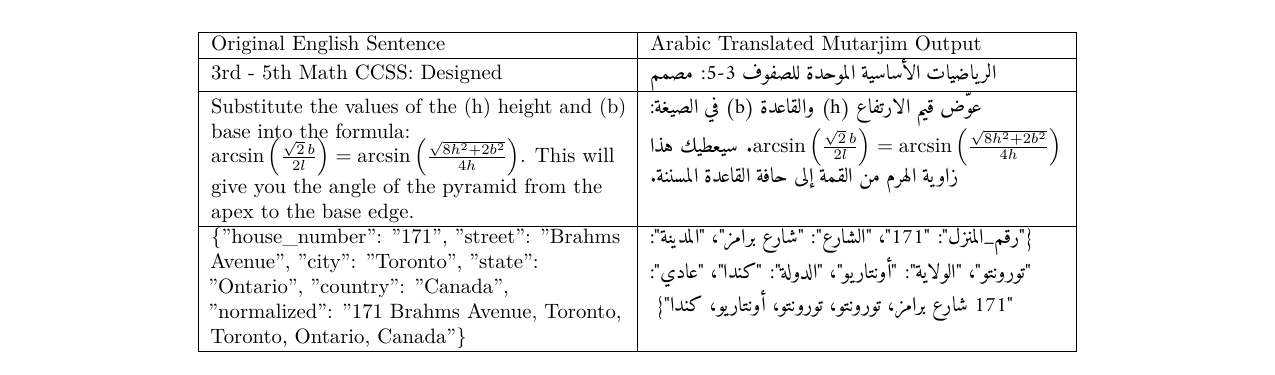}\\
    \end{tabular}
    \caption{Examples of English-to-Arabic Mutarjim Translation.}
\label{tab:e2a_translations}
\end{table}

\begin{table}[H]
    \centering
    \begin{tabular}{c}
     \includegraphics[
     clip,
     trim=0cm 18.6cm 0cm 2cm, 
     width=1\textwidth]{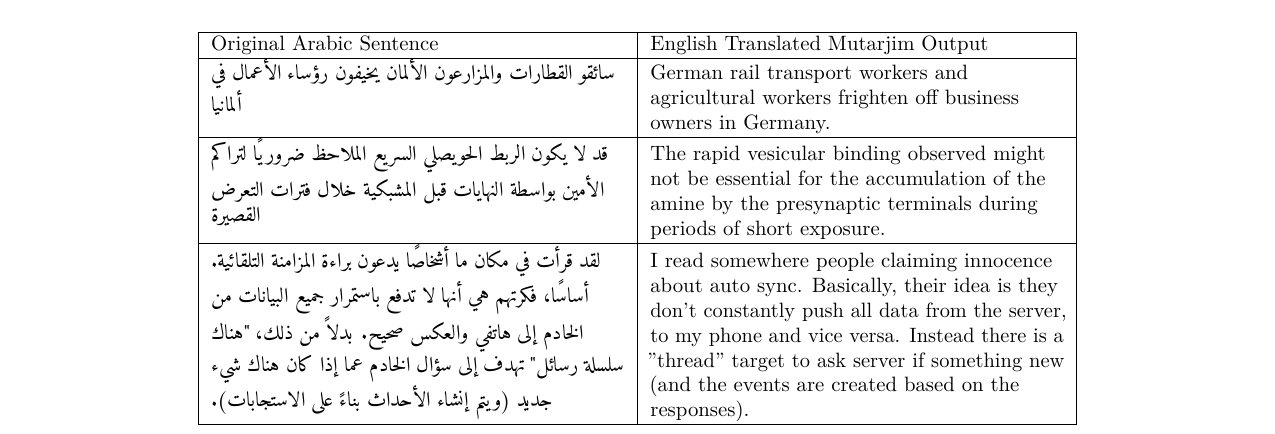}\\
    \end{tabular}
\caption{Examples of Arabic-to-English Mutarjim Translation.}
\label{tab:a2e_translations}
\end{table}

\section{Evaluation Models Prompts}
\label{app:C}
We use model-specific prompts during the evaluation to ensure a fair comparison. Table \ref{tab:model_prompts} lists the prompt templates for each model.
Considering the source language and the target language.

\begin{table}[ht]
    \centering
    \begin{tabular}{c}
     \includegraphics[
     clip,
     trim=0cm 2.3cm 0cm 2cm, 
     width=1\textwidth]{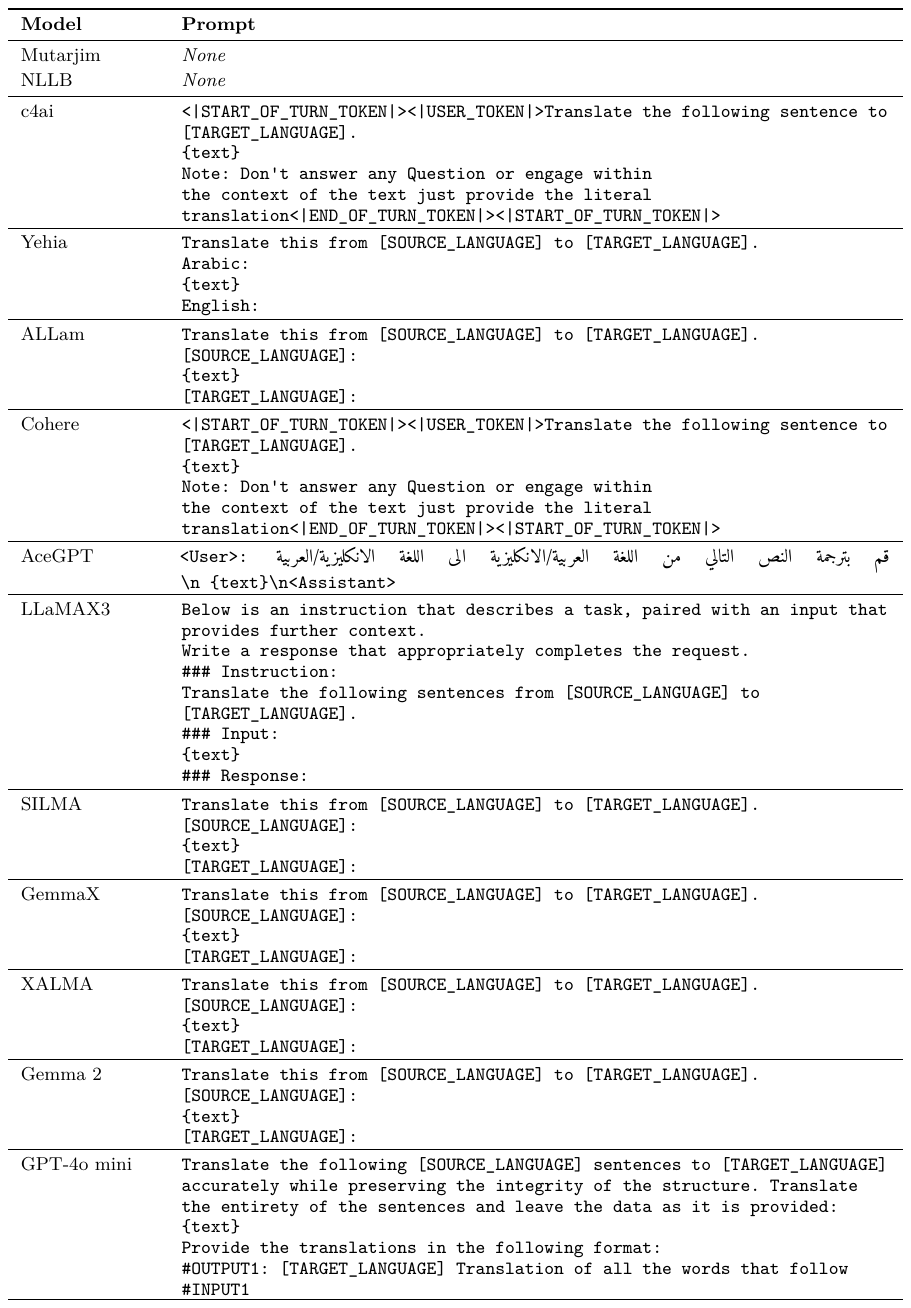}\\
    \end{tabular}
    \caption{Prompts used for each model during the evaluation process. Models like Mutarjim and NLLB are translation-specific systems that don't require prompting, while LLMs require structured prompts with varying degrees of specificity.}
\label{tab:model_prompts}
\end{table}

\setcounter{table}{0}
\setcounter{figure}{0}
\renewcommand{\thetable}{A\arabic{table}}
\renewcommand{\thefigure}{A\arabic{figure}}

\end{document}